# Psychological and Personality Profiles of Political Extremists

Meysam Alizadeh[1,2], Ingmar Weber[3], Claudio Cioffi-Revilla[2], Santo Fortunato[1], Michael Macy[4]

[1] Center for Complex Networks and Systems Research, School of Informatics and Computing, Indiana University, Bloomington, IN 47405, USA
[2] Computational Social Science Program, Department of Computational and Data Sciences, George Mason University, Fairfax, VA 22030, USA
[3] Qatar Computing Research Institute, Doha, Qatar
[4] Social Dynamics Laboratory, Cornell University, Ithaca, NY 14853, USA

## Abstract

Global recruitment into radical Islamic movements has spurred renewed interest in the appeal of political extremism. Is the appeal a rational response to material conditions or is it the expression of psychological and personality disorders associated with aggressive behavior, intolerance, conspiratorial imagination, and paranoia? Empirical answers using surveys have been limited by lack of access to extremist groups, while field studies have lacked psychological measures and failed to compare extremists with contrast groups. We revisit the debate over the appeal of extremism in the U.S. context by comparing publicly available Twitter messages written by over 355,000 political extremist followers with messages written by non-extremist U.S. users. Analysis of text-based psychological indicators supports the moral foundation theory which identifies emotion as a critical factor in determining political orientation of individuals. Extremist followers also differ from others in four of the Big Five personality traits.

## Introduction

Longstanding theoretical debates about the causes of extremism have recently resurfaced over efforts to explain recruitment into radical Islamic movements. Does the appeal of extremist groups reflect rational responses to material conditions in the diaspora or psychological and personality disorders associated with aggressive behavior, intolerance, conspiratorial imagination, and paranoia?

This question taps into competing psychological explanations for involvement in extremist political groups and social movements. The collective behavior school [1], drawing on postwar critical theory [2-3], focuses on psychological and personality disorders associated with susceptibility to radical beliefs and activities, regardless of economic and social privilege. Public choice [4] and resource mobilization theorists [5] distinguish between violent and non-violent extremism, noting similarities between the latter and mainstream political activity, both of which can be viewed as a rational response to perceived grievances. Simply put, non-violent extremists and those in the political mainstream do not differ psychologically, they differ in their material circumstances and access to political resources. More recently, moral psychologists have argued that the left differs from the right in the emotional resonance with different "moral foundations" [6-8]. The moral foundation theory asserts that moral judgments are grounded in emotion, not in moral reasoning [8], a theory with roots in David Hume's philosophical treatise on emotion as the underpinning of human morality [9]. Therefore, the two major hypotheses to be tested are:

1. *H1 (Collective Behavior Hypothesis): Extremists differ psychologically from mainstream activists regardless of ideology (whether left or right wing).*
2. *H2 (Moral Foundations Hypothesis): Left- and Right-wing activists differ psychologically from each other (whether extremist or mainstream)*

Research on the susceptibility to extremism has relied on surveys and ethnographic studies. Surveys have been administered to random samples as well as convenience samples (e.g. college students) to identify those whose responses fall at the extremes of the liberal-conservative spectrum [10-12] and who believe that the realization of fundamental social and economic changes requires militant action outside the electoral process. An immediate problem is that elicited responses to an interviewer are not equivalent to voluntary expressions of support for, agreement with, and endorsement of extremist groups and activities, including those that are non-violent. Other studies have used field observation of extremist groups [13-14], but these lack psychological measures and fail to compare political extremists with two important contrast groups: those that are political but non-extremist and those that are not political or extremist. Computational studies also suffer from lack of appropriate data for validation [e.g. 15-18].

Widespread use of social media by extremist organizations and their followers provides researchers with unprecedented opportunities to study the personality and psychological profiles of those who are susceptible to extremist appeals [e.g. 19-21]. Twitter messages have been shown to reveal underlying psychological and personality attributes that are reflected in word usages, a method pioneered by James Pennebaker [22]. We analyzed Twitter messages written by 355,000 American followers of non-violent U.S. extremist organizations and compared text-based indicators of personality attributes and psychological variables with results for random users, followers of apolitical celebrities, followers of far-left and far-right U.S. politicians, and followers of political moderate organizations. If followers of extremists differ from random users but are similar to followers of apolitical celebrities, that would indicate that the observed differences are due to being celebrity followers on Twitter, and not to being extremist. If followers of extremists differ from other users but are similar to elected officials with radical views, that would indicate that the observed differences are associated with holding extreme policy views, not to militancy or advocacy of extreme actions. If followers of extremists differ from other users but are similar to followers of moderate organizations working on the same issues (i.e. those who pursue similar goals but through mainstream channels), that would indicate that the observed differences are associated with the goal or ideology itself, not to being extremist on that goal or ideology.

*Extremists.* After defining ideologies in each of the left- and right-wing extremism categories (Supplementary Table 1), we collected up to 3200 available Twitter messages from each of over 355,000 followers of 23 independently identified left-wing (LWE) and 47 right-wing (RWE) extremist groups and organizations (see Materials and Methods section for definition of LWE and RWE and see Supplementary Table 2 and 3 for list of ideologies in each category). We classified the followers into individuals and organizations (see Supplementary Information for classifying users) and compared 149,693 right-wing and 27,927 left-wing *individual* extremist followers with 49,344 randomly chosen individual users (Supplementary Table 4) and 7,500 followers of five randomly chosen apolitical celebrities. Extremists can be distinguished from random users along multiple dimensions, including support for *action* outside mainstream electoral politics, and support for *policies* on the far right and far left such as a complete ban on abortion, deportation of immigrants, and support for radical environmental reform. We therefore also compared extremist followers with 6,000 followers of Bernie Sanders, Elizabeth Warren, Ted Cruz, and Donald Trump (Supplementary Table 5), 17,226 followers of anti-abortionist accounts, 17,191 followers of anti-immigrant accounts, 14,608 environmentalist followers (see Materials and Methods and Supplementary Table 6 for details of identifying moderate single-issue groups). We focus the analysis on the *followers* of extremist organizations, not the organizations themselves, in order to address the psychological appeal of extremist views, not the psychological expression.

*Affective indicators.* We measured affect by counting the words in all Twitter messages of a given user that appear in five psychological processes word lists of the Linguistics Inquiry and Word Count (LIWC) lexicon [22]: negative emotion (negemo), positive emotion (posemo), anger, anxiety, and certainty.

*Personality.* We used the IBM Watson Personality Insights service [23] to infer the Big Five personality traits [24] from the entire corpus of text written by each user. The Big Five personality traits include openness, conscientiousness, extraversion, agreeableness and neuroticism and they have become a standard classification system in psychological analysis [25]. Our methodology for inferring these traits has been independently validated by comparing text-based scores for 256 Twitter users with those same user's scores on the Big Five inventory [26].

## Results

Fig. 1 compares the mean affective profiles of left-wing and right-wing extremist followers with random users, followers of celebrities, and followers of prominent left- (i.e. Sanders and Warren) and right-wing (i.e. Trump and Cruz) politicians. We combined the followers of Sanders and Warren, and Cruz and Trump and report the average value of their psychological indicators as Sanders-Warren and Trump-Cruz in Fig. 1 respectively. Because followers of Sanders and Warren overlap much more so than those for Cruz and Trump, we tested whether the psychological indicators of Cruz and Trump followers were comparable before merging them (See supplementary Fig. 1). Results show that, compared to random users, left-wing extremist followers show less positive ($t = 26.51$, $p < 0.001$) and more negative emotion ($t = -26.76$, $p < 0.001$) but do not differ on the other three measures. Right-wing extremist followers show more negative emotion ($t = -25.03$, $p < 0.001$) but do not differ from random users on the other four measures ($t = 0.45$, $p = 0.65$). Negative affect does not distinguish the left and right, either for followers of extremists or radical politicians. Both LWE and RWE have higher negative affect than random users and celebrity followers but are similar to followers of Sanders, Warren, Cruz, and Trump. The results suggest similar negative emotional appeal of radical positions, regardless of the ideological direction, and whether or not the positions are associated with militant or mainstream political activity. This is not the case for positive emotion, for which we find large differences between RWE and LWE. Overall, the pattern indicates that negative emotion is associated with radical positions, whether left or right or extremist or mainstream, while positive emotion is lower for the left than the right, for the followers of politicians as well as extremists, but more so for the latter.

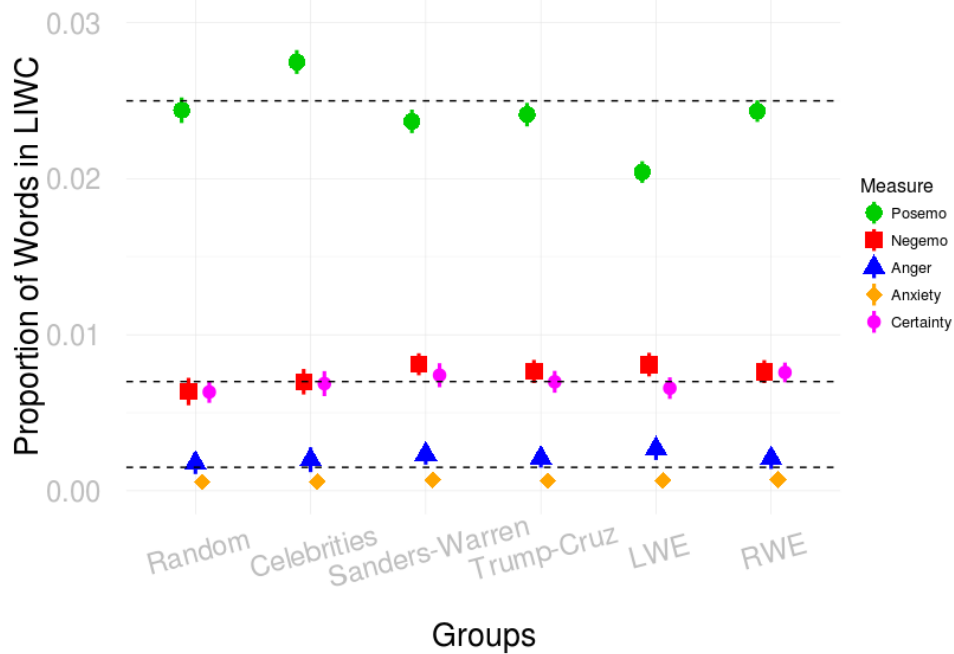

**Fig. 1 Emotional valence of extremist followers and non-extremist U.S. Twitter users.** The vertical axis is the group mean for the proportion of each user's total words (including multiple usages) that appear in the LIWC lexicon, along with 95% confidence intervals. The dotted lines do not indicate anything and are drawn for readers to better compare the values. Posemo (green circles) is shorthand for positive emotion and Negemo (red squares) is negative emotion. Results show that, compared to random users, left-wing extremist followers show less positive ($p < 0.001$) and more negative emotion ($p < 0.001$) but do not differ on the other three measures. Right-wing extremist followers show more negative emotion ($p < 0.001$) but do not differ on the other four measures. Extremist followers resemble the followers of politicians more so than the followers of celebrities.

The proportion of words that appear in the LIWC dictionaries, as reported in Fig. 1, is not in itself an informative metric. A more meaningful benchmark is the LIWC score among individuals that differ in emotional and mental health. Previous applications of LIWC dictionaries support the significance of this approach to influence the mental health research [27-31]. Harman et al [31] report median differences in Posemo, Negemo, Anger, and Anxiety for four clinical conditions: post traumatic stress disorder (PTSD), seasonal affective disorder (SAD), chronic depression, and bipolar disorder. The differences with random individuals we report for left-wing and right-wing extremists are as large or larger than the differences between random individuals and each of these four clinical disorders, and in the same direction (Supplementary Fig 2). This does not imply that extremists are mentally ill, but it does indicate that the effect sizes we report are comparable to those found for people with clinical disorders.

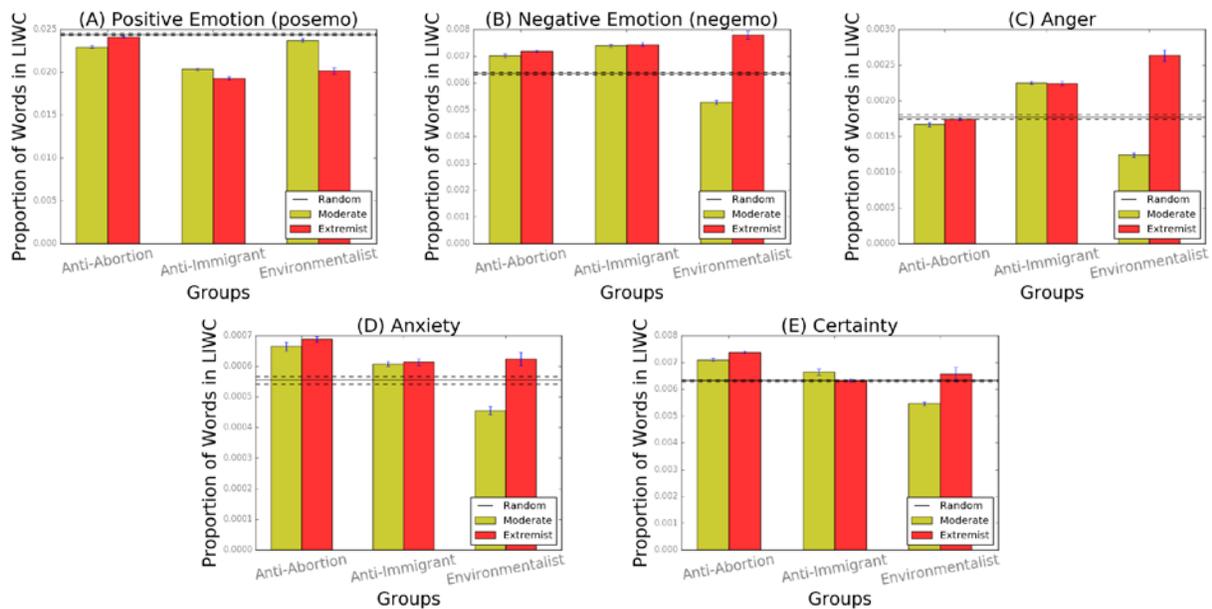

**Fig. 2.** Comparing psychological profiles of the followers of moderate and extremist single-issue groups, compared to random users. The vertical axis is the group mean for the proportion of each user's total words that appear in the LIWC lexicon, along with 95% confidence intervals. The vertical line in each plot represents the corresponding psychological measure value for random users. The two dashed lines show 95% confidence intervals. Among abortion opponents, followers of extremists show more negative emotion ($p < 0.01$), anxiety ($p < 0.001$), and certainty ($p < 0.01$) compared to moderates but do not differ in positive emotion or anger. Among those hostile to immigrants, followers of extremists show less positive emotion ($p < 0.001$) but otherwise do not differ from moderates. Among environmentalists, followers of extremists show less positive ($p < 0.05$) and more negative ($p < 0.001$) emotion, anger ($p < 0.001$), and anxiety ($p < 0.001$) compared to moderates. Overall, the differences in psychological profile between followers of extremist and moderate groups is much larger for left-wing extremists (environmentalists) than right-wing (anti-abortion and anti-immigrant).

Fig. 2 provides a more detailed view by focusing on followers of moderate and extremist issue-specific groups. The figure also reports scores for random users as a baseline. The results do not show a consistent pattern of emotional differences between politically engaged followers (whether moderate or extremist) and random users. Across all five measures, the emotional differences between moderates and extremist followers are consistently larger among environmentalist followers (on the left) than opponents of abortion and immigrants (on the right). Followers of extremist and moderate right-wing groups (whether on abortion or immigration) have higher negative affect and anxiety compared to random users. Followers of extremist environmentalist groups differ from moderate followers and random users in having less positive affect and more negative affect, anger, and certainty.

Overall, the results suggest that the psychological profile of extremist followers depends more on ideology (left vs right) than being extremist on that ideology. Among abortion opponents, followers of extremists show more positive emotion ($t = -5.31, p < 0.001$) and certainty ($t = -3.6, p < 0.001$) compared to moderates, but do not differ in negative emotion ($t = -1.93, p < 0.05$), anger ($t = -1.95, p < 0.05$), or anxiety ($t = -1.36, p < 0.1$). Among those hostile to immigrants, followers of extremists show less certainty ($t = 2.26, p < 0.05$) and positive emotion ($t = 4.36, p < 0.001$) but otherwise do not differ from moderates ($p < 0.1$). Among environmentalists, followers of extremists show less positive ($t = 8.33, p < 0.001$) and more negative ($t = -14.73, p < 0.001$) emotion, anger ($t = -15.99, p < 0.001$), certainty ($t = -4.13, p < 0.001$), and anxiety ($t = -6.76, p < 0.001$) compared to moderates. Overall, the differences in psychological profile between followers of extremist and moderate groups is much larger for left-wing extremists (environmentalists) than right-wing (anti-abortion and anti-immigrant).

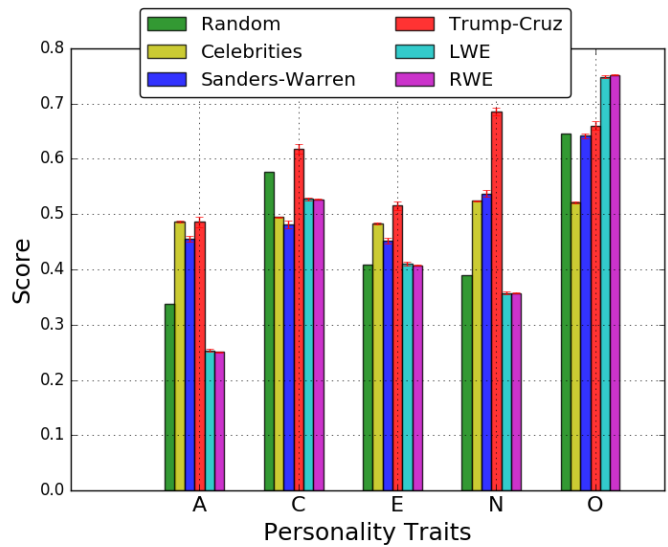

**Fig. 3.** Big Five Personality Profiles. The vertical axis is the mean score using IBM Watson Personality Insights for Agreeable, Conscientious, Extrovert, Neurotic, and Open. Results show that extremist followers (whether left or right) are less agreeable, less neurotic, and more open than non-extremists.

Fig. 3 extends the analysis to Big Five personality profiles as measured by the IBM Watson Personality Insights. The results show little difference between LWE and RWE on any of the Big Five traits, while extremist followers (whether left or right) are less agreeable, less neurotic, and more open than non-extremists. Compared to random Twitter users, both LWE and RWE are less agreeable ($t = -24.41, p < 0.001$) and ($t = -42.82, p < 0.001$), conscientious ($t = -16.06, p < 0.001$) and ($t = -26.54, p < 0.001$), and neurotic ($t = -12.65, p < 0.001$) and ($t = -19.26, p < 0.001$), more open ($t = 34.92, p < 0.001$) and ($t = 56.06, p < 0.001$), and similar in extraversion. While extremist followers differ in personality from random users, they do not differ from political moderates with similar ideology ($p < 0.001$, Supplementary Fig 3). The results further show that extremist followers differ in all five personality traits compared to the celebrity followers, and extreme politicians.

## Discussion

Psychological language analysis of Tweets by U.S. users shows that followers of extremist groups differ affectively from random Twitter users. We observed relatively large differences in positive and negative affect between the followers of left- and right-wing accounts, whether of extremist groups or politicians. A plausible interpretation is that those who seek wholesale changes in social and economic conditions tend to express themselves with greater affect. Followers of right-wing extremists are more prone to use positive emotion words, while LWE followers tend to express negative emotions. Analysis of single-issue groups also revealed much larger differences between extremist and moderate followers among left-wing environments than among right-wing opponents of abortion and immigrants.

These ideological differences in emotional profiles suggest that the affective differences reflect the ideological direction of extremists, not their militancy. The results are consistent with the moral foundation theory that differences between left- and right-wing individuals are associated with emotional resonance with different moral foundations, not to moral reasoning (for other empirical support of the moral foundation theory see Haidt and Graham [7] and Fulgoni et al [32]). The theory implies that the distinctive psychological profiles of LWE and RWE and can be attributed to their different empathic responses to different policy issues.

However, we found almost no ideological differences between LWE and RWE on any of the Big Five personality measures. Whether left or right, we found that extremist followers are less agreeable, less neurotic, and more open than non-extremists. The moral foundations theory appears to apply more to emotional measures than to personality, which is consistent with the theoretical predictions. In contrast, personality appears to be more closely associated with militancy than with the ideological direction to which militancy is applied. Indeed, less agreeable/more open extremists on both ends create seeds for political polarization - as they are impervious to the opinions of others and eager to disseminate their own.

Our results provide no evidence of a causal relationship between emotions and extremism. Those who are more emotional may be more susceptible to extremist appeals or those with extremist views may express themselves with more emotion, or both. Nor do these results rule out the alternative theory that material deprivation and political oppression encourage extremist views and emotional agitation.

We also have no evidence about extremist or violent behavior. Our analysis is focused on those who follow non-violent extremists and is therefore limited to the psychological profiles associated with attraction to extremist views. We do not have data for the followers of violent extremist groups or behavioral measures of their activities offline.

The results are nevertheless useful in providing evidence about the psychological profiles of those who are attracted to extremist views, as a possible precursor for violent behavior [33]. Although we do not find psychological attributes that consistently differentiate followers of militant extremists from followers of radical politicians, we cannot rule out the possibility that larger differences might characterize those who follow groups that advocate the use of violence or those who are members of those groups, not just followers on social media.

## Materials and Methods

According to the U.S. Department of Homeland Security [34], LWE refers to "a movement of groups or individuals that embraces anti-capitalist, Communist, or Socialist doctrines and seeks to bring about change through violent revolution rather than through established political processes. The term also refers to left-wing, single-issue extremist movements that are dedicated to causes such as environmentalism, opposition to war, and the rights of animals." RWE is defined as "a movement of rightwing groups or individuals who can be broadly divided into those who are primarily hate-oriented, and those who are mainly antigovernment and reject federal authority in favor of state or local authority. This term also may refer to rightwing extremist movements that are dedicated to a single issue, such as opposition to abortion or immigration."

We identified 47 right wing extremist groups listed as "hate groups" by the Southern Poverty Law Center [35]. For left wing extremists (LWE), we used Wikipedia to identify 23 groups that were described as socialist [36], communist [37], radical environmentalist [38], and anarchist [39]. We then identified 27,927 left-wing and 149,693 right-wing individual followers of these groups. We excluded violent extremists because there are too few cases to be found on Twitter. The National Consortium for the Study of Terrorism and Responses to Terrorism (START) has developed the Profiles of Perpetrators of Terrorism in the United States (PPT-US) database [40]. PPT-US includes detailed information on more than 140 organizations that have engaged in terrorist attacks in the US between 1970 and 2012. This list intersects with our list at the Aryan Nation and Ku Klux Klan as violent RWE, and Animal Liberation Front and Earth First! as violent LWE.

We identified 49,344 individual Twitter random users through the Followerwonk app [41] using the stratified sampling method based on the number of followers. That is, we divided the whole

range of 5 to 1000 number of followers to 20 segments (i.e. 5-50, 50-100, 100-150, …, 950-1000) and took 5000 followers uniformly at random from each segment. Then we classified the obtained accounts into individuals and organization accounts and took only individual ones.

We constructed psychological language profiles using the LIWC lexicon [42] which contains word lists for four affective processes – positive emotion, negative emotion, anger, and anxiety. We also include one words list from the cognitive processes category named certainty. Examples of affective processes words and word-phrases include "love", "nice", and "sweet" for positive emotion, "hurt", "ugly", and "nasty" for negative emotion, "worried" and "fearful" for anxiety, and "hate", "kill", and "annoyed" for anger. Cognitive processes words include "always" and "never" for certainty. We use these affective and cognitive processes word lists to count the number of usages across all the tweets in a user's Twitter "timeline" expressed as a proportion of the user's total word count. For a review of using Twitter data in health research see Sinnenberg et al [43].

In addition to affective profile, we also used the IBM Watson Personality Insights service [23] to infer the "Big Five" personality traits. This service analyzes text using LIWC lexicon [22]. The inferential algorithms were developed and validated at IBM by matching inferred scores with scores obtained from the same individuals using conventional survey instruments [44].

## Reference

1. Smelser NJ. Theory of collective behavior. Quid Pro Books; 2011 Aug 21.
2. Hoffer E. The True Believer: Thoughts on the Nature of Movements. New York NY: HarperCollins; 1951.
3. Adorno TW. Kulturkritik und Gesellschaft. InSoziologische Forschung in Unserer Zeit 1951 (pp. 228-240). VS Verlag für Sozialwissenschaften.
4. Berman E, Laitin D. Hard targets: Theory and evidence on suicide attacks. National Bureau of Economic Research; 2005 Nov 7.
5. McCarthy JD, Zald MN. The enduring vitality of the resource mobilization theory of social movements. In Handbook of sociological theory 2001 (pp. 533-565). Springer US.
6. Haidt J, Joseph C. Intuitive ethics: How innately prepared intuitions generate culturally variable virtues. Daedalus. 2004;133(4):55-66.
7. Haidt J, Graham J. When morality opposes justice: Conservatives have moral intuitions that liberals may not recognize. Social Justice Research. 2007 Mar 1;20(1):98-116.
8. Haidt J. The righteous mind: Why good people are divided by politics and religion. Vintage; 2012 Mar 13.
9. Hume D. A treatise of human nature. Courier Corporation; 2003.
10. Jost JT, Napier JL, Thorisdottir H, Gosling SD, Palfai TP, Ostafin B. Are needs to manage uncertainty and threat associated with political conservatism or ideological extremity?. Personality and social psychology bulletin. 2007 Jul;33(7):989-1007.

11. Kruglanski AW, Gelfand M, Gunaratna R. Terrorism as means to an end: How political violence bestows significance.
12. van Prooijen JW, Krouwel AP, Boiten M, Eendebak L. Fear Among the Extremes How Political Ideology Predicts Negative Emotions and Outgroup Derogation. Personality and social psychology bulletin. 2015 Feb 4:0146167215569706.
13. Canetti-Nisim D, Halperin E, Sharvit K, Hobfoll SE. A new stress-based model of political extremism personal exposure to terrorism, psychological distress, and exclusionist political attitudes. Journal of Conflict Resolution. 2009 Jun 1;53(3):363-89.
14. Atran S, Ginges J. Religious and sacred imperatives in human conflict. Science. 2012 May 18;336(6083):855-7.
15. Alizadeh M, Coman A, Lewis M, Cioffi-Revilla C. Intergroup conflict escalation leads to more extremism. Journal of Artificial Societies and Social Simulation. 2014;17(4);4.
16. Flache A, Macy MW. Small worlds and cultural polarization. The Journal of Mathematical Sociology. 2011 Jan 25;35(1-3):146-76.
17. Alizadeh M, Cioffi-Revilla C, Crooks A. THE EFFECT OF IN-GROUP FAVORITISM ON THE COLLECTIVE BEHAVIOR OF INDIVIDUALS'OPINIONS. Advances in Complex Systems. 2015 Feb 1;18(01n02):1550002.
18. Alizadeh M, Cioffi-Revilla C. Activation regimes in opinion dynamics: Comparing asynchronous updating schemes. Journal of Artificial Societies and Social Simulation. 2015 June 30;18(3);8.
19. Magdy W, Darwish K, Weber I. # FailedRevolutions: Using Twitter to study the antecedents of ISIS support. arXiv preprint arXiv:1503.02401. 2015 Mar 9.
20. Davidson T, Warmsley D, Macy M, Weber I. Automated Hate Speech Detection and the Problem of Offensive Language. arXiv preprint arXiv:1703.04009. 2017 Mar 11.
21. Ferrara E. Computational Social Science to Gauge Online Extremism. arXiv preprint arXiv:1701.08170. 2017 Jan 27.
22. Pennebaker JW, Francis ME, Booth RJ. Linguistic inquiry and word count: LIWC 2001. Mahway: Lawrence Erlbaum Associates. 2001;71(2001):2001.
23. IBM Corp., IBM Watson Personality Insights Service, 2016, Available from: http://www.ibm.com/smarterplanet/us/en/ibmwatson/developercloud/personality-insights.html, 2016.
24. Costa PT, McCrae RR. The revised neo personality inventory (neo-pi-r). The SAGE handbook of personality theory and assessment. 2008 Jun 24;2:179-98.
25. Goldberg LR, Johnson JA, Eber HW, Hogan R, Ashton MC, Cloninger CR, Gough HG. The international personality item pool and the future of public-domain personality measures. Journal of Research in personality. 2006 Feb 28;40(1):84-96.
26. Gou L, Zhou MX, Yang H. KnowMe and ShareMe: understanding automatically discovered personality traits from social media and user sharing preferences. In Proceedings of the 32nd annual ACM conference on Human factors in computing systems 2014 Apr 26 (pp. 955-964). ACM.

27. Chung C, Pennebaker JW. The psychological functions of function words. Social communication. 2007:343-59.
28. Park M, Cha C, Cha M. Depressive moods of users portrayed in Twitter. In Proceedings of the ACM SIGKDD Workshop on healthcare informatics (HI-KDD) 2012 Aug 12 (pp. 1-8).
29. De Choudhury M, Gamon M, Counts S, Horvitz E. Predicting Depression via Social Media. In ICWSM 2013 Jul 8 (p. 2).
30. Coppersmith G, Harman C, Dredze M. Measuring Post Traumatic Stress Disorder in Twitter. In ICWSM 2014 May 16.
31. Harman GC. Quantifying mental health signals in Twitter. ACL 2014. 2014 Jun 27;51.
32. Fulgoni D, Carpenter J, Ungar L, Preotiuc-Pietro D. An Empirical Exploration of Moral Foundations Theory in Partisan News Sources. In LREC 2016.
33. McCauley C, Moskalenko S. Mechanisms of political radicalization: Pathways toward terrorism. Terrorism and political violence. 2008 Jul 1;20(3):415-33.
34. Department of Homeland Security. Domestic Extremism Lexicon. 2009. Available in: http://www.webcitation.org/5gYPnOstE.
35. Southern Poverty Law Center, Extremist Files, 2016. Available in: https://www.splcenter.org/fighting-hate/extremist-files.
36. Wikipedia, Socialist Parties in the United States. 2017; 2, 19. Available in: http://en.wikipedia.org/wiki/Category:Socialist_parties_in_the_United_States.
37. Wikipedia, Communists Parties in the United States. 2017; 2, 19. Available in: http://en.wikipedia.org/wiki/Category:Communist_parties_in_the_United_States.
38. Wikipedia, Radical Environmentalism, 2017; 2, 19. Available in: http://en.wikipedia.org/wiki/Radical_environmentalism.
39. Wikipedia, List of Anarchist Organizations. 2017; 2, 19. Available in: http://en.wikipedia.org/wiki/List_of_anarchist_organizations.
40. Miller E, Smarick K. Profiles of Perpetrators of Terrorism in the United States (PPT-US).
41. Followerwonk, Twitter Analytics: Find, Analyze, and Optimize for Social Growth. 2016. Available in: https://moz.com/followerwonk/.
42. Pennebaker JW, Boyd RL, Jordan K, Blackburn K. The development and psychometric properties of LIWC2015. 2015 Sep 15.
43. Sinnenberg L, Buttenheim AM, Padrez K, Mancheno C, Ungar L, Merchant RM. Twitter as a Tool for Health Research: A Systematic Review. American journal of public health. 2017 Jan;107(1): e1-8.
44. Yarkoni T. Personality in 100,000 words: A large-scale analysis of personality and word use among bloggers. Journal of research in personality. 2010 Jun 30;44(3):363-73.

# Supplementary Information

## Identifying Political Extremist Ideologies

Categories are based on the U.S. Department of Homeland Security's (DHS) "Domestic Extremism Lexicon" (DHS 2009). But this lexicon did not categorize the extremists groups into the left/right wing divisions. Another organization that provides a list and profiles of the extremists in the US is the Southern Poverty Law Center (SPLC) which has a database of the U.S. extremist individuals and groups based on their ideology without categorizing them to left/right extremists. Here we combine the categories defined by the DHS and SPLC and categorize them into the broader left/right wing extremists based on the definitions provided by the DHS and publicly available sources such as Wikipedia. Detailed names of the groups in the Supplementary Information.

**Table S1.** List of the Left/Right Ideological Extremists Groups in the US.

| Left Wing Extremists Groups | Right Wing Extremists Groups |
|---|---|
| Anarchist Extremism | Anti-Abortion Extremism |
| Animal Rights Extremism | Anti-Immigration Extremism |
| Environmental Extremism | Anti-LGBT Extremism |
| Socialist/Communist | Anti-Muslim Extremism |
| | Black Separatist |
| | Christian Identity |
| | General Hate |
| | Holocaust Denial Extremism |
| | Neo-Nazis |
| | Patriot Movements |
| | Skinheads |
| | White Nationalist |

# Identifying Left- and Right-Wing Extremist Groups

Southern Poverty Law Center (SPLC) maintains the profiles of all hate groups in the US. According to the definition of the right-wing extremism given by the Department of Homeland Security, we extracted the following groups from SPLC's database who have active Twitter account. We consider an account as active if it contains at least one tweet in 2015 and have more than 10 tweets overall. We have also excluded the protected accounts in the following table. Finally, it should be noted that we have not included the local branches of a given ideology and only focused on the main group's Twitter account.

**Table S2.** List of Left-Wing Extremist Groups in US

| No. | Group | Category |
| --- | --- | --- |
| 1 | Democratic Socialists of America | Socialist/Anti-Capitalism |
| 2 | Boston Socialism | Socialist |
| 3 | Communist Party USA | Socialist |
| 4 | CrimethInc. | Anarchist |
| 5 | Deep Green Resistance | Radical Environmentalist/Feminist |
| 6 | Freedom Socialist Party | Socialist |
| 7 | International Action Center | Socialist |
| 8 | Kshama Sawant | Socialist |
| 9 | Liberty Union Party | Socialist/Anti-war |
| 10 | News and letters Committees | Communist |
| 11 | North American Animal Liberation Press Office | radical environmentalist |
| 12 | NYC ISO | Socialist |
| 13 | Party for Socialism and Liberation | Socialist |
| 14 | Peace and Freedom Party | Socialist/Feminist |
| 15 | Progressive Labor Party | Socialist |
| 16 | Radical Women | Socialist-Feminist |
| 17 | Socialist Action | Socialist |
| 18 | Socialist Alternative | Socialist |
| 19 | Socialist Equality Party | Socialist |
| 20 | Socialist Party USA | Socialist |
| 21 | Socialist Worker | Socialist |
| 22 | UMass Boston ISO | Socialist |
| 23 | Workers World Party | Communist |

**Table S3.** List of Right-Wing Extremist Groups in the US

| No. | Group Name | Ideological Type |
|---|---|---|
| 1 | American Family Association | anti-LGBT |
| 2 | American Freedom Defense Initiative | anti-muslim |
| 3 | American Life League | anti-abortion |
| 4 | American Nazi Party | Neo-Nazi |
| 5 | American Renaissance | White nationalist |
| 6 | American Vision | anti-LGBT |
| 7 | Americans for Truth About Homosexuality | anti-LGBT |
| 8 | Americans United for Life | anti-abortion |
| 9 | Aryan Brotherhood | Neo-Nazi |
| 10 | Bryan Fischer | anti-LGBT |
| 11 | Chalcedon Foundation | anti-LGBT |
| 12 | Christian Action Network | anti-muslim |
| 13 | Chuck Buldwin | Patriot Movement |
| 14 | Constitution Party | American nationalism |
| 15 | David Barton | anti-LGBT |
| 16 | David Irving | Holocaust Denial |
| 17 | Dove World Outreach Center | anti-LGBT |
| 18 | Faith Freedom International | anti-muslim |
| 19 | Faithful Word Baptist Church | anti-LGBT |
| 20 | Family Research Council | anti-LGBT |
| 21 | Federation for American Immigration Reform | anti-immigrant |
| 22 | Frank Gaffney | anti-muslim |
| 23 | Illinois Family Institute | anti-LGBT |
| 24 | John "Molotov" Mitchell | anti-LGBT |
| 25 | Joseph Farah | Patriot Movement |
| 26 | Lou Engle | anti-LGBT |
| 27 | Michael Hill | Neo-Confederate |
| 28 | National Policy Institute | White nationalist |
| 29 | National Right to Life Committee | anti-abortion |
| 30 | National Socialist Movement | Neo-Nazi |
| 31 | Nationalist Movement | White nationalist |
| 32 | New Black Panther Party | Black separatist |
| 33 | Public Advocate of the United States | anti-LGBT |
| 34 | Robert Spencer | anti-muslim |
| 35 | SaveCalifornia.com | anti-LGBT |
| 36 | Tea Party Nation | General Hate |
| 37 | The Political Cesspool | White nationalist |
| 38 | The Remnant/The Remnant Press | Radical traditional Catholicism |
| 39 | Tony Alamo Christian Ministries | General Hate |
| 40 | Tony Perkins | anti-LGBT |
| 41 | Traditional Values Coalition | anti-LGBT |
| 42 | United Families International | anti-LGBT |
| 43 | VDARE Foundation | White nationalist |
| 44 | Washington Summit Publishers | White nationalist |
| 45 | Westboro Baptist Church | anti-LGBT |
| 46 | WND news | Patriot Movement |
| 47 | World Congress of Families | anti-LGBT |

# Classifying Users

We use multiple-wave human coding on CrowdFlower to create a training set ($N = 14{,}500$) and feed it to *humanizr* (McCorriston et al 2015), a support-vector machine based classifier, to identify individual (vs. organizations) accounts in our data. We design a job with instructions and questions in the format illustrated below.

**Overview:**

We need your help to classify the type of given Twitter accounts and understand whether they support or tweet about a given ideology.

**Definitions:**

You will be asked about whether a user is a supporter of a given ideology. For that, you need to know the definition of all 16 ideologies covered in this job. At each question, if you click on the provided ideology name, you'll be directed to a Wikipedia page explaining that ideology. The list of all covered ideologies throughout this job is as following:

Anarchism, Radical Animal Rights, Anti-Abortion, Anti-Immigrant, Anti-LGBT, Anti-Muslim, Anti-War, Communism, General Hate, Neo-Confederate, Neo-Nazi, Patriot Movement, Radical Environmentalism, Radical Traditional Catholicism, Socialism, White Nationalism.

**Instructions:**

You will be asked to open Twitter profiles of different individuals and read their bio and last 30 tweets and ask the following questions:

(**A) User Type:**

Based on user's name, bio, and profile picture, please determine its type:

1. Individual: User name is an actual human name / their self-description mentions them as a person.
2. Organization: Government department, corporation, NGO, campaign, or news media.
3. Spammer: User promotes misleading and undesirable information, mostly of a commercial nature.
4. Page not accessible: Only if the Twitter page is deleted, suspended, OR the tweets are protected and the type cannot be identified based on name, profile picture, and bio.

*NOTE*: Even if the user's tweets are protected, in most cases you can identify its type based on bio, name, and profile picture. But if you couldn't identify the type based on name, profile picture, or bio, then select "Page Not Accessible"

**(B) Individual Sub-Type:**

After reading users' bio and last 30 tweets, you should determine whether the user is a celebrity, or someone belong to an organization that mostly tweets about her/his job, or just someone sharing her/his personal matters and opinions. Please note that in case of "Belong to organization", we are NOT asking whether the user has specified her/his job. Rather, we are looking for those whose tweets can be seen as their job-related issues. Thus, if the bio says that the user is a director, journalist, or policy analyst in some organization, but the tweets are personal, you should choose "Personal".

1. Celebrity: a star in entertainment, sports, etc.
2. Organization: a person whose tweets mostly represent her/his organizational role and job rather than personal issues and opinions, such as most of the journalists, business owners who tweet about their products, scientists who mostly tweet about the results of their research, policy analysts who mostly tweet share links to their organization, etc.
3. Personal: a person tweeting about their personal views on issues.
4. None of the above: if tweets are protected you MUST select this.

**(C) Supporter of a Given Ideology:**

Here, you should look for at least one word or hashtag or link or photo in user's bio or tweets/re-tweets that is DIRECTLY related to the given ideology. For example, having only this tweet "LGBT boast in their political power and finances, but we will boast in the name of the Lord" is enough for you to choose "Supporter" for the Anti-LGBT ideology. Or having #Socialist in the bio is also enough to choose "Supporter" for the "Socialism" Ideology. Here are the options for this part:

1. Supporter: There is at least one tweet or explicit word or hashtag in user's bio related to the given ideology.
2. Not Supporter: There is no related tweet or explicit hashtag or word in the bio.
3. Not Identifiable: ONLY if user's tweets are protected

*VERY IMPORTANT NOTE*: Please AVOID subjective judgments without any clue. For example, being a church, Pastor, or Christian conservative do not automatically make the user "Anti-LGBT", unless you find at least a word, hashtag, link, etc against LGBT in their overall profile. So if someone says she/he is a Christian conservative/Pastor/Patriot/etc, but you cannot find any word, hashtag, link, tweet, re-tweet, etc. related to anti-LGBT in her/his overall profile, then you should choose "Not Supporter".

Please visit XYZ's Twitter page and check their profile (e.g., location, bio, and profile picture) and read their 20-30 recent tweets.

[XYZ's Twitter page]

Which type is this Twitter account?

○ Individual (a real person)

○ Organization (e.g., news media or company)

○ Spammer (providing misleading and undesirable, commercial information)

○ Page not accessible (Only if the page has been deleted or suspended OR tweets are protected and the type cannot be identified)

Even if the user's tweets are protected, in most cases you can identify its type based on bio, name, and profile picture.

Which sub-type is this Twitter account?

○ Celebrity (a well-known figure from entertainment, sports, ...)

○ Belongs to organization (e.g., journalist, businessman, scientist, ...)

○ Personal (individual not obviously affiliated with an organization)

○ None of the above

Select 'Belongs to organization' only if user's tweets are mostly related to user's organization. Select 'None of the above' if tweets are protected.

Is this user a supporter of the following ideology?
[Patriot Movement](#)

○ Yes

○ No

○ Not identifiable (ONLY if user's tweets are protected)

Look for at least one explicit clue in user's overall profile. Do not make subjective judgment

## Data Summary

Our data collection process begins with finding the user IDs and profile information of the followers of our list of the ideological extremists groups. We only keep those IDs that their language is English, are from US, have between 10 and 1000 number of followers, and are not "verified". We impose the number of followers and "not verified" constraints to exclude possible journalists, researchers, groups and organizations, and any other popular accounts, as we only look for individual followers.

**Table S4.** Data Summary

| Categories | Left | Right | Total |
|---|---|---|---|
| Violent | 3,956 | 58 | 4,014 |
| Active | 3,938 | 29,484 | 33,422 |
| Non-Active | 12,643 | 83,159 | 95,802 |
| Protected | 2,507 | 18,298 | 20,805 |
| Total Unique Individual Followers | 23,044 | 130,999 | 154,043 |
| Organizations | 3,549 | 15,247 | 18,796 |
| Inaccessible | 395 | 2,508 | 2,903 |
| User following both left and right | 939 | 939 | 939 |
| Total Followers after Preprocessing with Unique IDs | 27,927 | 149,693 | 177,620 |
| Total Followers after Preprocessing | 37,883 | 224,393 | 262,276 |
| Total Followers with Unique IDs | 69,583 | 285,602 | 355,185 |
| Total Followers | 90,286 | 398,964 | 489,250 |

# Followers of Political and Non-political Accounts

After crawling the user IDs of the Twitter followers of the above celebrities, we uniformly sampled 10,000 followers at random. After excluding those users whose location is not US, language is not English, and are unverified, we used *humanizr* to classify the users into individuals and organization. Finally, we uniformly sampled 1,300 users at random from the pool of individual accounts.

**Table S5.** List of the Celebrities.

| Category | Celebrity Name | Celebrity's Twitter Screen Name | Followers Sample Size |
|---|---|---|---|
| Non-political | Ahuvah Berger-Burcat | @ahoova | 1,500 |
| | Oprah | @oprah | 1,500 |
| | Shaq | @shaq | 1,500 |
| | Taylor Swift | @taylorswift13 | 1,500 |
| | Tom Hanks | @tomhanks | 1,500 |
| Extreme Democrat | Bernie Sanders | @BernieSanders | 1,500 |
| | Elizabeth Warren | @elizabethforma | 1,500 |
| Extreme Republican | Ted Cruz | @tedcruz | 1,500 |
| | Donald Trump | @realDonaldTrump | 1,500 |

## Political Moderates Categories

We use the votesmart.org website to identify the non-extremist interested groups. For other ideologies listed in Table S1, we could not find corresponding non-extremist groups. In case of radical animal right ideology, our list of groups does not include any non-violent extremist group and only contains violent extremist animal rights groups.

**Table S6.** List of the political moderates with same ideologies groups.

| Name | Ideology |
| --- | --- |
| Democrats for the Life of America | Anti-Abortion |
| National Right to Life Committee | Anti-Abortion |
| Susan B Anthony List | Anti-Abortion |
| Americans for Legal Immigration | Anti-Immigration |
| Californians for Population Stabilization | Anti-Immigration |
| NumbersUSA | Anti-Immigration |
| Earth Policy Institute | Environmentalist |
| Environmental Working Group | Environmentalist |
| Natural Resources Defense Council | Environmentalist |
| Sierra Club | Environmentalist |

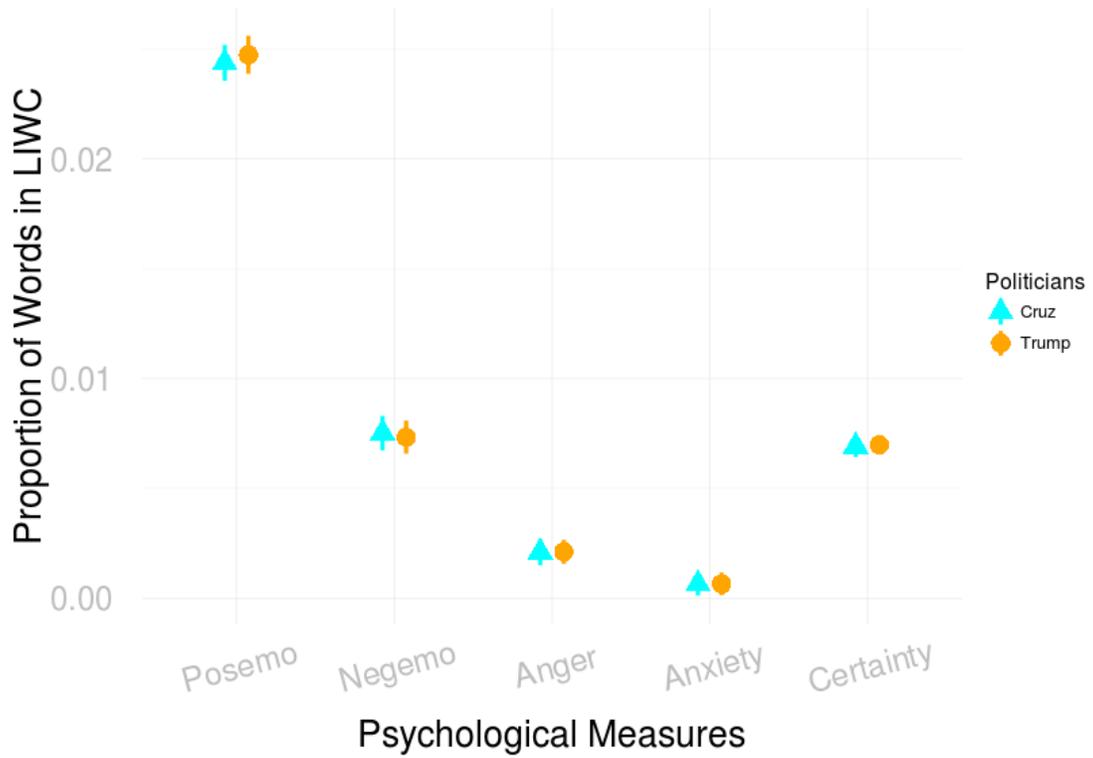

**Fig S1: Comparing linguistic psychological indicators of the followers of Donald Trump and Ted Cruz.** The vertical axis is the group mean for the proportion of each user's total words (including multiple usages) that appear in the LIWC lexicon, along with 95% confidence intervals. Posemo refers to positive emotion and Negemo is negative emotion. Results show that followers of Trump and Cruz share similar psychological indicators and therefore we combined them.

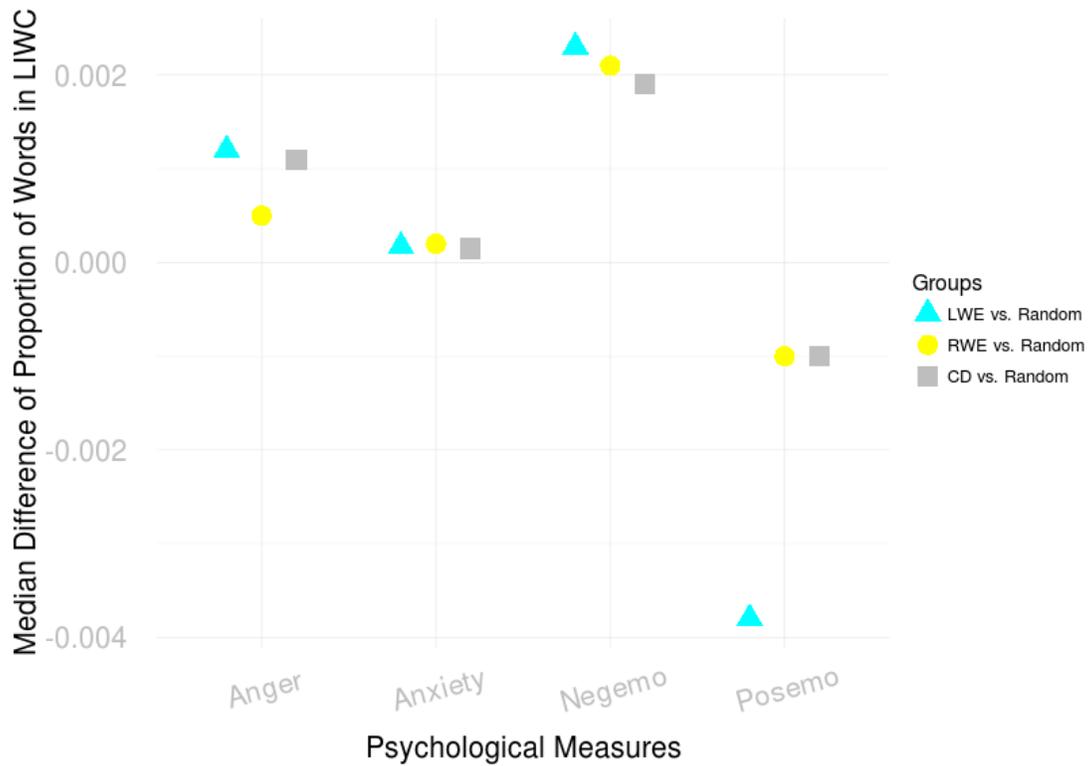

**Fig S2: Comparing median differences of random users, extremist followers, and individuals with clinical disorder (CD)**. The shown values for people with CD is the average median values for four clinical conditions including post-traumatic stress disorder (PTSD), seasonal affective disorder (SAD), chronic depression, and bipolar disorder. The differences with random individuals we report for left-wing and right-wing extremists are as large or larger than the differences between random individuals and each of these four clinical disorders, and in the same direction. This does not imply that extremists are mentally ill, but it does indicate that the effect sizes we report are comparable to those found for people with clinical disorders.

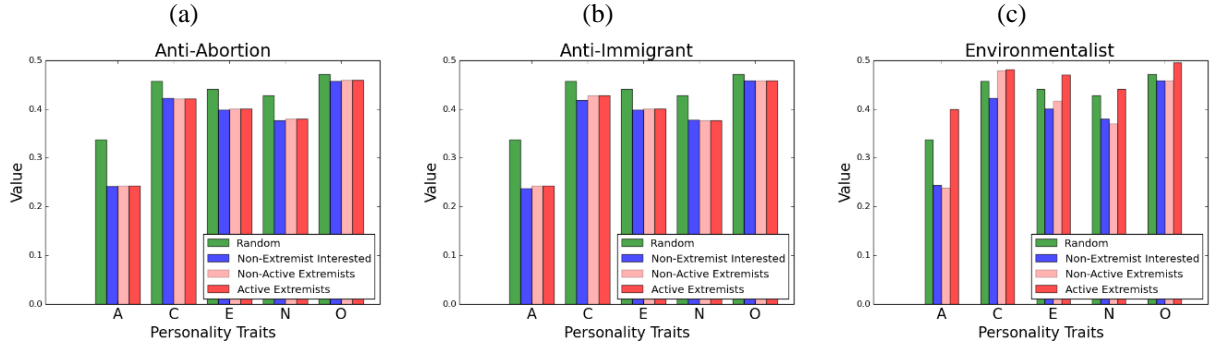

**Fig S3: Comparing Personality Traits of Extremists with Non-Extremists with Similar Goals and Random Individuals**. **a)** Comparing extremist and non-extremist individuals who hold anti-abortion ideology. While there is a significant difference between an average person and anti-abortion extremists/non-extremists ($p < 0.001$), there is no difference between extremists and non-extremists across the big five personality traits ($p < 0.01$). **b)** Comparing extremist and non-extremist individuals who hold anti-immigration ideology. The random set of individuals significantly differ from extremists and non-extremists ($p < 0.001$). However, the differences between extremists and non-extremists anti-immigrants are very little and none significant ($p < 0.01$). **c)** Comparing extremist and non-extremist individuals who hold extreme environmentalism ideology. All differences between random individuals and extreme/non-extreme environmentalists are significant ($p < 0.001$). There exist some differences between non-extreme and extremists across the big five personality traits but they are statistically significant at $p < 0.05$.

# Reference

McCorriston, James, David Jurgens, and Derek Ruths. "Organizations are Users Too: Characterizing and Detecting the Presence of Organizations on Twitter." *Ninth International AAAI Conference on Web and Social Media*. 2015.